\documentclass[twoside]{IEEEtran}

\usepackage{algorithm}
\usepackage{algpseudocode}
\usepackage{amsmath}
\usepackage{graphicx}
\usepackage{fancyhdr}
\usepackage{multirow}
\usepackage{threeparttable}

\begin{document}

\title{Adaptive Chemical Reaction Optimization for Global Numerical Optimization}

\author{James J.Q. Yu, \IEEEmembership{Student Member,~IEEE},
        Albert Y.S. Lam, \IEEEmembership{Member,~IEEE,}
        and Victor O.K. Li, \IEEEmembership{Fellow,~IEEE}
}

\maketitle
\pagestyle{empty}
\thispagestyle{plain}
% insert page header and footer here for IEEE PDF Compliant
\fancypagestyle{plain}{
	\fancyhf{}      % clear all header and footer fields
	\fancyfoot[L]{978-1-4799-7492-4/15/\$31.00~\copyright2015~IEEE}
	\fancyfoot[C]{}
	\fancyfoot[R]{}
	\renewcommand{\headrulewidth}{0pt}
	\renewcommand{\footrulewidth}{0pt}
}

\pagestyle{fancy}{
	\fancyhf{}
	\fancyfoot[R]{}}
\renewcommand{\headrulewidth}{0pt}
\renewcommand{\footrulewidth}{0pt}

\begin{abstract}
A newly proposed chemical-reaction-inspired metaheurisic, Chemical Reaction Optimization (CRO), has been applied to many optimization problems in both discrete and continuous domains. To alleviate the effort in tuning parameters, this paper reduces the number of optimization parameters in canonical CRO and develops an adaptive scheme to evolve them. Our proposed Adaptive CRO (ACRO) adapts better to different optimization problems. We perform simulations with ACRO on a widely-used benchmark of continuous problems. The simulation results show that ACRO has superior performance over canonical CRO.
\end{abstract}

\begin{IEEEkeywords}
Chemical Reaction Optimization, continuous optimization, adaptive scheme, metaheuristic, evolutionary algorithm.
\end{IEEEkeywords}

\section{Introduction}

Optimization techniques are frequently used in science and engineering research and development. For example, the problem of computing the contact force in mechanics can be regarded as a quadratic programming problem \cite{NaceuraGuobBatozc2004BlankOptimizationIn}. In economics, firms are usually expected to maximize their profit, while customers aim to maximize their utilities. So the asset price can be modelled by optimization techniques \cite{TapiaCoello2007ApplicationsMultiobjective}. We also encounter numerous optimization problems in our daily lives, such as finding the shortest route to a destination while avoiding traffic congestion \cite{YangLi2010HybridGeneticAlgorithm}, and assigning various tasks to different time slots to maximize the amount of work done before the deadline \cite{ChiuHsuYeh2006GeneticAlgorithmReliability}.

Chemical Reaction Optimization (CRO), proposed in \cite{LamLi2010ChemicalReactionInspired}, is a chemical-reaction-inspired general-propose metaheuristic. CRO is variable population-based and it was initially designed to solve combinatorial problems. Then Lam \textit{et al.} proposed a Real-Coded CRO (RCCRO) \cite{LamLiYu2012RealCodedChemical}, which aims to solve continuous optimization problems. In this paper, with abuse of notation, we use CRO to refer to both CRO in \cite{LamLi2010ChemicalReactionInspired} and RCCRO in \cite{LamLiYu2012RealCodedChemical}. In a chemical reaction, the reactants tend to release the excessive energy to the environment to approach relatively lower energy states. CRO utilizes this tendency and incorporates the chemical reaction ideas to construct an optimization algorithm.

CRO has been applied to solve many optimization problems. Lam and Li adopted CRO to solve Quadratic Assignment Problem (QAP), Resource-Constraint Project Scheduling Problem, and Channel Assignment Problem in \cite{LamLi2010ChemicalReactionInspired}. The first two are classical combinatorial problems while the last one is a practical NP-hard combinatorial problem according to \cite{SahniGonzalez1976Pcompleteapproaximation}. They are used as benchmark problems for many other algorithms \cite{DigalakisMargaritis2000BenchmarkFunctionsGenetic}\cite{Taillard1991Robusttaboosearch} and CRO outperforms many of such algorithms \cite{LamLi2010ChemicalReactionInspired}. Xu \textit{et al.} \cite{XuLamLi2011ChemicalReactionOptimization} applied CRO to a task scheduling problem in grid computing, which is a multi-objective NP-hard optimization problem. Lam \textit{et al.} addressed the cognitive radio spectrum allocation problem with CRO in \cite{LamLi2010ChemicalReactionOptimization}. Yu \textit{et al.} proposed a sensor deployment problem for air pollution detection and solved it with CRO \cite{YuLiLam2012SensorDeploymentAir}. All these problems mentioned are NP-hard combinatorial optimization problems and CRO was shown to be effective and efficient in solving these problems. Lam \textit{et al.} also studied the convergence characteristics of CRO in \cite{LamLiXu2013ConvergenceChemicalReaction}.

Lam \textit{et al.} then proposed RCCRO to tackle continuous problems \cite{LamLiYu2012RealCodedChemical}. They made three major modifications on the structures of molecule and algorithm to make CRO applicable to continuous problems. A classic benchmark set with 23 continuous functions were adopted for testing and the simulation results indicated that RCCRO outperformed most of the other Evolutionary Algorithms (EAs) compared. Lam \textit{et al.} proposed a population transition problem for peer-to-peer live streaming and applied CRO to solve the problem \cite{LamXuLi2010ChemicalReactionOptimization}. Yu \textit{et al.} proposed a CRO-based Artificial Neural Network (CROANN) algorithm to train neural networks (NN) for classification \cite{YuLamLi2011EvolutionaryArtificialNeural}. CROANN also outperformed most previously proposed EA-based NN training algorithms. Yu \textit{et al.} also studied different perturbation functions for RCCRO in \cite{YuLamLi2012RealcodedChemical}. They compared the Gaussian distribution function used in \cite{LamLiYu2012RealCodedChemical} with three other distribution functions for the 23 benchmark functions and provided guidelines for further RCCRO research.

According to our experience, proper parameter setting is key for CRO to solve optimization problems efficiently. However, we usually tune the parameters according to our experiences and by trial and error. Moreover, there are no theoretical results on parameter setting for EAs in general. To avoid the time-consuming parameter tuning process, we may pursue adaptive approaches. We developed a deterministic parameter control scheme for CRO in \cite{LamLiYu2012RealCodedChemical}, but this scheme is still lacking as it only modifies the \textit{StepSize} parameter of CRO deterministically and there is no feedback to inform the modification. On the contrary, adaptive parameter schemes have been proposed for some other EAs and they allow one to minimize the effort on parameter tuning and focus on the formulation of the optimization problem and the design of the algorithm operators.

This paper is focused on developing an adaptive scheme for CRO. This new version of CRO with the adaptive scheme is called Adaptive CRO (ACRO). ACRO reduces the number of parameters defined in canonical CRO from eight to three, and makes them adaptive to feedback information in the course of searching. To evaluate its optimization performance, we use ACRO to solve a set of benchmark functions mainly selected from the latest IEEE CEC Real-Parameter Single Objective Optimization Competition problem set \cite{LiangQuSuganthanHernandez-Diaz2013ProblemDefinitionsand}.

The rest of this paper is organized as follows. In Section \ref{sec:CRO}, we briefly introduce the framework of CRO. We present ACRO with parameter adaptation in Section \ref{sec:ACRO}, and Section \ref{sec:experiment} describes the benchmark problems used to examine the performance of ACRO and the simulation configuration. The comparisons with the canonical CRO are presented in Section \ref{sec:result}. We conclude the paper in Section \ref{sec:conclusion}.

\section{Chemical Reaction Optimization} \label{sec:CRO}

In this section, we introduce the general framework of CRO. We consider a number of molecules in a container, with an attached central energy buffer.

\subsection{Molecule}
A molecule is the basic operating agent of CRO. CRO controls and manipulates a set of molecules systematically to explore the solution space. Each molecule is characterized by attributes such as the molecular structure, potential energy, and kinetic energy. We relate chemical reactions to optimization by assigning mathematical meanings to these attributes. In a typical implementation of CRO, we include the following attributes:

\subsubsection{Molecular Structure}
We denote the molecular structure by $\omega$, which represents the solution carried by the molecule. The detailed structure of $\omega$ is problem-dependent, which means that the configuration of $\omega$ is usually determined by the encoding scheme of the problem. For different optimization problems, $\omega$ can be a vector, a matrix, a graph, or a string \cite{LamLi2012ChemicalReactionOptimization:}.

\subsubsection{Potential Energy}
We use $\textit{PE}_\omega$ to denote the potential energy (PE) of the molecule with molecular structure $\omega$. In CRO, $\textit{PE}_\omega$ represents the objective function value of the solution carried by the molecule, i.e., $f(\omega)$. So our goal of optimization is to reduce $\textit{PE}_\omega$ as much as possible. Since ``Every reacting system seeks to achieve a minimum of free energy" \cite{Masel2001ChemicalKinetics&}, the chemical reaction process drives the molecules toward the lowest energy state on the potential energy surface (PES).

\subsubsection{Kinetic Energy} \label{subsub:KE}
We use $\textit{KE}_\omega$ to denote the kinetic energy (KE) of the molecule with molecular structure $\omega$. In CRO, $\textit{KE}_\omega$ is regarded as the tolerance of the molecule to accept a worse $\omega^\prime$ with higher $\textit{PE}_{\omega^\prime}$ than the existing $\textit{PE}_\omega$. The main purpose of introducing KE to CRO is to avoid the algorithm from getting stuck in local optima. This will be elaborated in Section \ref{subsec:elementaryreactions}.

\subsection{Energy Conservation Law}
Energy transformation between different forms and conservation of energy are features which distinguish CRO from other metaheuristics \cite{LamLiYu2012RealCodedChemical}. The law of conservation of energy states that although energy is allowed to transform between types, the total energy in an isolated system (i.e. the container in CRO) shall remain constant. We use $\textit{E}_{\textit{total}}$ and $\textit{E}_{\textit{total}}^{\prime}$ to denote the total energy in the container before and after an elementary reaction, respectively. Similarly, we denote the energy in the central energy buffer before and after an elementary reaction by $\textit{E}_{\textit{buffer}}$ and $\textit{E}_{\textit{buffer}}^{\prime}$, respectively. In an elementary reaction, consider that the original molecules $\omega_1, \omega_2, ..., \omega_n$ are transformed into new molecules $\omega_1^{\prime}, \omega_2^{\prime}, ..., \omega_p^{\prime}$. The total energy before the change is given by
\begin{equation}
\begin{aligned}
\textit{E}_{\textit{total}}=&\textit{E}_{\textit{buffer}}+\textit{PE}_{\omega_1}
+\textit{KE}_{\omega_1}+\textit{PE}_{\omega_2}+\textit{KE}_{\omega_2}+...\\
&+\textit{PE}_{\omega_n}+\textit{KE}_{\omega_n},
\end{aligned}
\end{equation}
and it will remain constant, i.e.,
\begin{equation}
\begin{aligned}
\textit{E}_{\textit{total}}^{\prime}=&\textit{E}_{\textit{buffer}}^{\prime}
+\textit{PE}_{\omega_1^{\prime}}+\textit{KE}_{\omega_1^{\prime}}
+\textit{PE}_{\omega_2^{\prime}}+\textit{KE}_{\omega_2^{\prime}}+...\\
&+\textit{PE}_{\omega_p^{\prime}}+\textit{KE}_{\omega_p^{\prime}} \\
=&\textit{E}_{\textit{total}}.
\end{aligned}
\end{equation}

The transformation among $\textit{E}_\textit{buffer}$, \textit{PE}, and \textit{KE} for CRO is stated in \cite{LamLi2010ChemicalReactionInspired} and interested readers can refer to \cite{LamLi2012ChemicalReactionOptimization:} for more details.

\begin{table*}
	\caption{Classification of Elementary Reactions}
	\label{tbl:classreaction}
	\small
	\begin{center}
		\begin{tabular}{r|l|l}
			\hline
			& Uni-Molecular Collision & Inter-Molecular Collision \\ \hline
			Constant Population Size Collision & On-wall Ineffective Collision & Inter-Molecular Ineffective Collision \\ \hline
			Variable Population Size Collision & Decomposition &  Synthesis\\ \hline
		\end{tabular} 
	\end{center}
\end{table*}

\subsection{Elementary Reactions} \label{subsec:elementaryreactions}
There are four types of elementary reactions defined in CRO, including on-wall ineffective collision, decomposition, inter-molecular ineffective collision, and synthesis. We classify the elementary reactions by the number of molecules involved (see Table \ref{tbl:classreaction}). For uni-molecular reactions, only one molecule is involved in the reaction as the reactant. For inter-molecular collisions, two molecules act as the reactants of the reaction. An elementary reaction may or may not change the total number of molecules in the system. Constant population size collisions do not change the number of molecules in the container, while variable population size collisions increase or decrease the number of molecules by one. A uni-molecular reaction is triggered when a molecule collides with a wall of the container, while an inter-molecular reaction corresponds to two molecules colliding with each other. Moreover, only successful reactions (subject to energy constraints) will modify the molecular structures of the involved molecules. CRO modifies the molecular structures of the molecules to explore the solution space with these elementary reactions. The detailed information about these elementary reactions can be found in \cite{LamLi2010ChemicalReactionInspired} and \cite{LamLi2012ChemicalReactionOptimization:}.

\section{ACRO} \label{sec:ACRO}

\begin{table*}
	\caption{CRO Parameter Categories}
	\label{tbl:parameter}
	\small
	\begin{center}
		\begin{tabular}{r|l}
			\hline
			Category & Parameter \\ \hline
			\multirow{3}*{Energy Related} & \textit{iniKE} \\ & \textit{iniBuffer} \\ & \textit{LossRate} \\ \hline
			\multirow{2}*{Reaction Related} & \textit{DecThres} \\ & \textit{SynThres} \\ \hline
			Real-Code Related & \textit{StepSize} \\ \hline
		\end{tabular} 
	\end{center}
\end{table*}

In this section we propose our new CRO with parameter adaptation.

Canonical CRO has eight parameters, namely, initial population size (\textit{iniPopSize}), initial molecular kinetic energy (\textit{iniKE}), initial central energy buffer (\textit{iniBuffer}), molecular collision rate (\textit{CollRate}), energy loss rate (\textit{LossRate}),  decomposition occurrence threshold (\textit{DecThres}), synthesis occurrence threshold (\textit{SynThres}), and perturbation step size (\textit{StepSize}). These parameters cooperate to control the performance of the algorithm.

In our proposed new CRO algorithm, we modify six parameters of CRO, including \textit{iniKE}, \textit{iniBuffer}, \textit{LossRate}, \textit{DecThres}, \textit{SynThres}, and \textit{StepSize}. We categorize them into three classes: energy-related, reaction-related, and real-coded-related as shown in Table \ref{tbl:parameter}. We will discuss our modifications in the following subsections. In general, we will simplify the design of \textit{iniKE}, \textit{iniBuffer}, and \textit{LossRate}. Then we will use a new parameter to replace both \textit{DecThres} and \textit{SynThres}. The new parameter \textit{ChangeRate}, along with a newly proposed elementary reaction selection scheme, will make the occurrence of elementary reactions adaptive. Finally we will employ a modified ``1/5 success rule" to make \textit{StepSize} adaptive. The remaining parameters (\textit{iniPopSize} and \textit{CollRate}) retain the original definitions in CRO presented in \cite{LamLi2010ChemicalReactionInspired}.

\subsection{Energy-Related Parameters}
Parameters \textit{iniKE}, \textit{iniBuffer}, and \textit{LossRate} cooperate with each other to control the convergence of CRO \cite{LamLiYu2012RealCodedChemical}. \textit{iniKE} and \textit{iniBuffer} determine the total energy in the system before the algorithm starts, while \textit{LossRate} controls the rate of transforming KE to the central energy buffer in on-wall ineffective collisions.

In CRO, KE controls the tolerance of the molecule to accept a worse structure \cite{LamLi2010ChemicalReactionInspired}. So a large KE may potentially lead to a situation in which the molecule keeps accepting worse solution structures, resulting in bad optimization performance. Meanwhile, small KE may cause the molecule to get stuck in local optima since the molecule does not have enough tolerance to jump out of local optima. As stated in \cite{LamLi2010ChemicalReactionInspired}, the energy of the central buffer and PE in molecules can be transformed into KE, and this process is partially controlled by \textit{LossRate} in on-wall ineffective collision. So \textit{iniKE}, \textit{iniBuffer}, and \textit{LossRate} also impact the performance, and proper values of these parameters are very important in controlling the convergence of the algorithm.

\subsubsection{iniKE}
This parameter is used to control the initial KE of molecules. When the algorithm starts and the initial population is generated, all the molecules in the population are assigned with KE according to \textit{iniKE}.  The selection of \textit{iniKE} shall be made based on the characteristics of the objective function value.

In ACRO, we first generate a population of molecules $\textit{pop}_\textit{init}=\{m_1, m_2, ..., m_\textit{iniPopSize}\}$ by choosing feasible solutions in the solution space randomly. Their corresponding PEs are determined. We label the molecular structures of the two molecules with the largest and the smallest PE by $\omega_l$ and $\omega_s$ in $\textit{pop}_\textit{init}$, respectively. Then we define the initial kinetic energy of all the molecules as
\begin{equation}
\textit{iniKE}=(\textit{PE}_{\omega_l}-\textit{PE}_{\omega_s})\times \textit{iniPopSize}
\end{equation}
 and this initial KE rule is applied whenever a new molecule is generated in subsequent decompositions.

\subsubsection{iniBuffer}
The central energy buffer $\textit{E}_{\textit{buffer}}$ accumulates energy in on-wall ineffective collisions, and energy is withdrawn in decomposition \cite{LamLi2010ChemicalReactionInspired}. As $\textit{E}_{\textit{buffer}}$ is initiated according to \textit{iniBuffer}, this value is very important in determining whether a decomposition is successful, especially in the early stage of the search.

Decomposition is often manipulated with operators performing vigorous modifications to the molecular structures, assisting molecules to jump out of local optima. But a decomposition that occurs too early may potentially hamper the convergence of the algorithm, especially in those cases in which the decomposition modifies the molecules that are still evolving, i.e. not stuck in a local optimum. This hinders ACRO from performing an effective optimization search as the molecule will switch to another region of PES instead of performing a local search. In other words, this may lead to missing a possible global optimum in a particular region. As stated previously, the success rate of decomposition is partially controlled by $\textit{E}_{\textit{buffer}}$. We can reduce the initial energy in the central buffer to suppress the decomposition at the early stage since a small central buffer, initialized with a small \textit{iniBuffer}, renders it less likely for decomposition to occur \cite{LamLiYu2012RealCodedChemical}. When the molecules have collided and explored the solution space for some time, $\textit{E}_{\textit{buffer}}$ is then gradually increased due to on-wall ineffective collisions, and the molecules are more likely to get stuck in the local optima. Since $\textit{E}_{\textit{buffer}}$ is getting larger, more decomposition will be successful and the molecules can explore other parts of the solution space without getting stuck.

Therefore, according to the previous analysis, we assign zero to \textit{iniBuffer} to suppress decomposition in the early stage of searching in ACRO. The idea of setting the initial buffer to zero was also supported numerically by the simulations in \cite{XuLamLi2011ChemicalReactionOptimization} and \cite{LamLiYu2012RealCodedChemical}.

\subsubsection{LossRate}
Parameter \textit{LossRate} is used in the energy transformation process of the on-wall ineffective collision. When a molecule is involved in a successful on-wall ineffective collision, a portion of the energy that the molecule originally holds will be transferred to $\textit{E}_{\textit{buffer}}$.

In a successful on-wall ineffective collision, the excessive energy $\textit{E}_{\textit{excess}}$ is divided into two parts. One part becomes the KE of the new molecule according to
\begin{equation}
\textit{KE}_{\omega^\prime}=\textit{E}_{\textit{excess}}\times\textit{q}
\end{equation}
where $\omega^\prime$ is the new molecular structure, and \textit{q} is a random number generated from the interval $[\textit{LossRate}, 1]$. The remaining energy $\textit{E}_{\textit{excess}}-\textit{KE}_{\omega^\prime}$ is transferred to $\textit{E}_{\textit{buffer}}$. Hence, \textit{LossRate} affects the KE a molecule possesses after a successful on-wall ineffective collision.

In many previous works of CRO, \textit{LossRate} is set to be a very small number (e.g., 0.1 in \cite{LamLiYu2012RealCodedChemical}, 0.2 in \cite{YuLamLi2012RealcodedChemical}). In our ACRO algorithm, the parameter \textit{LossRate} is designed to be a molecular attribute, instead of the original global attribute. The value for \textit{LossRate} of each molecule in the system is approximated by a modified folded normal distribution. This distribution is generated from a normal distribution with a mean value of 0 and standard deviation of 0.3. All negative values in this normal distribution is ``folded'' over by taking their absolute values, and all values larger than 1 are considered as 1. By doing so, we attempt to make the energy loss rate a small enough number, but without losing the original searching characteristics provided by large \textit{LossRate}.

\subsection{Reaction-Related Parameters}
In CRO, we use \textit{CollRate}, \textit{DecThres}, and \textit{SynThres} to control the ratio among the four kinds of elementary reactions. We first apply \textit{CollRate} to determine whether a uni-molecular collision or an inter-molecular collision would happen in the current iteration. Then one or two molecules are randomly chosen from the population as the input molecule(s) to an elementary reaction. For a uni-molecular collision, we compare the ``inactive degree" \cite{LamLi2010ChemicalReactionInspired} of the molecule with \textit{DecThres}. If the inactive degree is larger than \textit{DecThres}, decomposition will occur. For an inter-molecular collision, the KEs of the two input molecules are compared with \textit{SynThres}. If both of the KEs are smaller than \textit{SynThres}, synthesis will occur. However, this scheme may result in an imbalance between the number of decompositions and syntheses when inappropriate values are set to \textit{CollRate}, \textit{DecThres}, and \textit{SynThres}. For example, if we have too many syntheses, the population size is likely to be reduced to one most of the time during simulation. Then it is highly likely for the molecules to get stuck in local optima and the population diversity is no longer maintained. On the other hand, if there are many decompositions, too many new molecules will be generated. The limited number of function evaluations will be exhausted without thoroughly searching the potential regions. Then, CRO may turn into a random search. For simplicity but without loss of generality, we assume that the population diversity can be maintained when the number of molecules is stable in the course of searching, with population size similar to the initial value, i.e., \textit{iniPopSize}.

In ACRO, we introduce a new parameter \textit{ChangeRate} to replace the original parameters \textit{DecThres} and \textit{SynThres} to control the frequency of decompositions and syntheses. Generally \textit{ChangeRate} describes the probability of having decomposition and synthesis. For example, if we set $\textit{ChangeRate}=0.01$, then we will roughly have $1000\times0.01=10$ decompositions or syntheses among 1000 elementary reactions. However, \textit{ChangeRate} alone is not enough to control the population size and maintain the population diversity. As in the previous example, it is possible that all 10 reactions are syntheses. Then if the \textit{iniPopSize} is set to 11, the population size will finally be reduced to one. In such a case, the population diversity is still not maintained. In order to handle this problem, we shall increase the occurrence rate of decomposition when the population size is smaller than \textit{iniPopSize}, and increase the rate of synthesis when the population size is large. To do this, we introduce two population feedbacks $\textit{f}_\textit{dec}$ and $\textit{f}_\textit{syn}$ to control the occurrences of decomposition and synthesis, respectively.

In ACRO, in each iteration, the algorithm will first decide whether it is a variable population-size collision or constant population-size collision using \textit{ChangeRate}. Then the algorithm will decide whether a decomposition or synthesis shall happen. We define the population feedback term as
\begin{equation}
\textit{f}_\textit{pop}=\frac{\textit{curPopSize}-\textit{iniPopSize}}{\textit{iniPopSize}}.
\end{equation}
Then the decomposition and synthesis feedbacks are defined as
\begin{equation}
\textit{f}_\textit{dec}=\frac{1}{2}\times(1-\textit{f}_\textit{pop})
\end{equation}
and
\begin{equation}
\textit{f}_\textit{syn}=1-\textit{f}_\textit{dec}=\frac{1}{2}\times(1+\textit{f}_\textit{pop}),
\end{equation}
respectively, where \textit{curPopSize} is the current population size. $\textit{f}_\textit{dec}$ and $\textit{f}_\textit{syn}$ determine the probability the current iteration is a decomposition and a synthesis, respectively.

The population feedback term $\textit{f}_\textit{pop}$ is positive when \textit{curPopSize} is larger than \textit{iniPopSize}. A positive $\textit{f}_\textit{pop}$ will encourage the occurrence of synthesis by increasing its probability and decreasing the probability of decomposition, and vice versa. If the current population size equals \textit{iniPopSize}, $\textit{f}_\textit{pop}=0$ and $\textit{f}_\textit{dec}=\textit{f}_\textit{syn}=0.5$, which means that decomposition and synthesis are equally likely to occur. If the current population size is one and \textit{iniPopSize} is not one, $\textit{f}_\textit{pop}=-1$ and the probability of having synthesis is zero. This matches the requirement of synthesis that at least two molecules in the population are needed. If \textit{iniPopSize} and \textit{curPopSize} are both one, despite $\textit{f}_\textit{syn}=0.5$, we stipulate that only decomposition can occur. To conclude, $\textit{f}_\textit{dec}$ cooperates with $\textit{f}_\textit{syn}$ to control the population size and keeps its fluctuations within a small range by controlling the relative occurrences of decomposition and synthesis.

\subsection{Real-Code-Related Parameter}
For continuous optimization problems, there are infinite feasible values for each variable. In CRO, we apply the zero-mean Gaussian distribution in the neighborhood searching operator for continuous optimization problems. Parameter \textit{StepSize} is employed to control the variance of the Gaussian distribution, which determines ``how far" the molecule can go in the on-wall and inter-molecular ineffective collisions \cite{LamLiYu2012RealCodedChemical}. This parameter is set to be a constant in CRO, and decreased at a predefined rate in the adaptive scheme \cite{LamLiYu2012RealCodedChemical}.

In CRO, the neighborhood search operator works as follows. Given a molecular structure, we first randomly pick an element from it. Then we randomly generate a new number $\textit{s}_\Delta\sim\mathcal{N}(0,\textit{StepSize})$ according to the Gaussian distribution and add this number to the value of the element. Suppose the selected element is the i-th element in the solution, and its original value is $\textit{s}_i$. Then the new value ${\textit{s}_i}^\prime$ of this element after applying the neighborhood search operator is
\begin{equation}
{\textit{s}_i}^\prime=\textit{s}_i+\textit{s}_\Delta .
\end{equation}

In CRO, a proper value of \textit{StepSize} is critical to the performance of the algorithm. While an overly large \textit{StepSize} prevents the algorithm from converging, an unreasonably small \textit{StepSize} will result in the algorithm getting stuck in local optima. The modification in \textit{StepSize} is twofold: initial value assignment and its subsequent update.

\subsubsection{Initial Value of StepSize} \label{sss:initStepSize}
In ACRO, we set an initial value to the \textit{StepSize} and then update this parameter in the course of searching. For each element in a solution, there is an upper bound and lower bound to confine the feasible range. We set the initial \textit{StepSize} for this element to be
\begin{equation}
\textit{StepSize}_{\textit{init},\textit{i}}=(\textit{Bound}_{\textit{upper},\textit{i}}-\textit{Bound}_{\textit{lower},\textit{i}})/2
\end{equation}
where $\textit{StepSize}_{\textit{init},\textit{i}}$ is the initial \textit{StepSize} for the i-th element in the solution, $\textit{Bound}_{\textit{upper},\textit{i}}$ and $\textit{Bound}_{\textit{lower}}$ are the upper and lower bounds for the i-th element. For simplicity, we omit \textit{i} in the subscript of \textit{StepSize} in the sequel.

\subsubsection{Evolution of StepSize}
We adapt the ``1/5 success rule" \cite{Schwefel1995EvolutionandOptimum} to modify \textit{StepSize} in the course of searching. This rule was originally stated as follows:

\textit{After every n mutations, check how many successes have occurred over the preceding 10n mutations. If the number is less than 2n, multiply the step lengths by the factor 0.85; divide them by 0.85 if more than 2n successes occurred.}

The ``1/5 success rule" has been applied to EP, ES and other EAs successfully \cite{MullerSchraudolphKoumoutsakos2002Stepsizeadaptation} and demonstrated outstanding performance. So we adopt this rule to manipulate the ``stepsize'' parameter adaptively. However it cannot be employed directly in CRO since there is no ``mutation" defined in CRO. In CRO, whenever an elementary reaction is successfully performed, new solution(s) are generated, and we have an ``update" process at the end of each iteration \cite{LamLi2010ChemicalReactionInspired}. In ``update", the algorithm compares the newly explored solutions with the best ever solution. If one of the new solutions is even better, the algorithm stores its PE (objective function value) and regards this solution as the best ever solution. In ACRO, we apply a modified ``1/5 success rule" in ``update".

The modification is as follows. The frequency at which \textit{StepSize} is changed is kept the same, i.e., $10n$ updates, and the adaptation factor of \textit{StepSize} is also kept constant at 0.85. \textit{n} is set to be one hundredth of the maximum allowable function evaluation number (\textit{FELimit}). The new update mechanism with ``1/5 success rule" is presented in Algorithm \ref{alg:stepsize}.

\begin{algorithm}
\caption{\sc{OnUpdateNewSolution($\omega_{new}$, $\omega_{best}$)}}
	\begin{algorithmic}[1]
	\State Compare the new solution $\omega_\textit{new}$ with the stored best ever solution $\omega_\textit{best}$.
	\If {$\omega_\textit{new}$ is better}
		\State $\omega_\textit{best}=\omega_\textit{new}$
		\State This update is successful.
	\EndIf
	\For {Every \textit{n} updates}
		\If {Previous \textit{10n} updates have more than \textit{2n} successful ones}
			\State $\textit{StepSize}=\textit{StepSize}/0.85$
		\Else
			\State $\textit{StepSize}=\textit{StepSize}\times0.85$
		\EndIf
	\EndFor
	\end{algorithmic}
	\label{alg:stepsize}
\end{algorithm}

\begin{table*}
	\caption{Parameter Modifications in  ACRO}
	\label{tbl:reducepara}
	\begin{center}
		\begin{tabular}{l|l|l}
			\hline
			Parameter & \multicolumn{1}{c|}{Initial Design in CRO} & \multicolumn{1}{c}{Modification in ACRO}\\\hline
			\multirow{2}*{\textit{iniKE}} & & The initial KE is assigned according to the PE of the initial population.\\
			& & The formula is $\textit{iniKE}=(\textit{PE}_{\omega_l}-\textit{PE}_{\omega_s})\times \textit{iniPopSize}$. \\\cline{1-1}\cline{3-3}
			\textit{iniBuffer} & & The initial central energy buffer is set to zero.\\\cline{1-1}\cline{3-3}
			\multirow{2}*{\textit{LossRate}} & & This parameter is unique for each molecule, and the value is generated\\
			& User assigns a proper initial value. & from a folded normal distribution.\\\cline{1-1}\cline{3-3}
			\textit{DecThres} & & These two parameters are replaced with a new parameter \textit{ChangeRate},\\
			\textit{SynThres} & & to control the occurrences of decomposition and synthesis.\\\cline{1-1}\cline{3-3}
			\multirow{2}*{\textit{StepSize}} & & The initial \textit{StepSize} is set according to the feasible solution bounds. It\\
			& & evolves according to a modified ``1/5 success rule".\\\hline
			\textit{iniPopSize} & \multicolumn{2}{c}{\multirow{2}*{No modification}}\\
			\textit{CollRate} & \multicolumn{2}{c}{} \\\hline
		\end{tabular}
	\end{center}
\end{table*}

\subsection{Summary}
In ACRO, we make \textit{StepSize} adaptive. We remove \textit{LossRate}, \textit{DecThres}, and \textit{SynThres} and introduce a new parameter \textit{ChangeRate} to control the occurrence of the elementary reactions. In addition, we formulate rules to assign initial values for \textit{iniKE} and \textit{iniBuffer}. So we reduce the total number of parameters from eight to three as indicated in Table \ref{tbl:reducepara}. The decrease in the total number of parameters can reduce the effort to tune the parameter values. We will show that ACRO performs equally well or even better than the canonical CRO in the next section.

\section{Experiment Setting} \label{sec:experiment}

\begin{table*}
	\caption{Benchmark Functions}
	\tiny
	\label{tbl:benchmark}
	\begin{center}
		\begin{threeparttable}
			\tiny
			\begin{tabular}{lllll}
				\hline
				$f$ & Function & Solution Transformation\tnote{*} & Name \\
				\hline
				$f_1$ & $\begin{aligned}\sum\nolimits^D_{i=1}z^2_i\end{aligned}$ & $\mathbf{z} = \mathbf{x} - \mathbf{o}$& Shifted Sphere Function\\
				$f_2$ & $\begin{aligned}\sum\nolimits^D_{i=1}\sum\nolimits^i_{j=1}z^2_i\end{aligned}$ & $\mathbf{z} = \mathbf{x} - \mathbf{o}$& Shifted Schwefel's Problem 1.2 \\
				$f_3$ & $\begin{aligned}\sum\nolimits^D_{i=1}\sum\nolimits^i_{j=1}z^2_i\end{aligned}$ & $\mathbf{z} = \mathbf{M}(\mathbf{x} - \mathbf{o})$& Shifted Rotated Schwefel's Problem 1.2 \\
				$f_4$ & $\begin{aligned}max\{|z_i|,1\leq i\leq D\}\end{aligned}$ & $\mathbf{z} = \mathbf{x} - \mathbf{o}$ & Shifted Schwefel's Problem 2.21 \\
				$f_5$ & $\begin{aligned}max\{|z_i|,1\leq i\leq D\}\end{aligned}$ & $\mathbf{z} = \mathbf{M}(\mathbf{x} - \mathbf{o})$ & Shifted Rotated Schwefel's Problem 2.21 \\
				$f_6$ & $\begin{aligned}\sum\nolimits^D_{i=1}|z_i|+\prod\nolimits^D_{i=1}|z_i|\end{aligned}$ & $\mathbf{z} = (\mathbf{x} - \mathbf{o}) \times 0.1$ & Shifted Schwefel's Problem 2.22 \\
				$f_7$ & $\begin{aligned}\sum\nolimits^D_{i=1}|z_i|+\prod\nolimits^D_{i=1}|z_i|\end{aligned}$ & $\mathbf{z} = \mathbf{M}(\mathbf{x} - \mathbf{o}) \times 0.1$ & Shifted Rotated Schwefel's Problem 2.22 \\
				$f_8$ & $\begin{aligned}\sum\nolimits^{D-1}_{i=1}(100(z^2_i-z_{i+1})^2 + (z_i-1)^2)\end{aligned}$ & $\mathbf{z} = (\mathbf{x} - \mathbf{o}) \times 0.3$ & Shifted Rosenbrock's Function \\
				$f_9$ & $\begin{aligned}\sum\nolimits^{D-1}_{i=1}(100(z^2_i-z_{i+1})^2 + (z_i-1)^2)\end{aligned}$ & $\mathbf{z} = \mathbf{M}(\mathbf{x} - \mathbf{o}) \times 0.3$ & Shifted Rotated Rosenbrock's Function \\
				$f_{10}$ & $\begin{aligned}10^6z^2_1+\sum\nolimits^D_{i=2}z^2_i\end{aligned}$ & $\mathbf{z} = \mathbf{x} - \mathbf{o}$ & Shifted Discus Function \\
				$f_{11}$ & $\begin{aligned}-20\exp(-0.2\sqrt{\frac{1}{n}\sum\nolimits^n_{i=1}z^2_i}) - \exp(\frac{1}{n}\sum\nolimits^n_{i=1}\cos 2\pi z_i)\\ + 20 + e\end{aligned}$ & $\mathbf{z} = (\mathbf{x} - \mathbf{o}) \times 0.32$ & Shifted Ackley's Function \\
				$f_{12}$ & $\begin{aligned}-20\exp(-0.2\sqrt{\frac{1}{n}\sum\nolimits^n_{i=1}z^2_i}) - \exp(\frac{1}{n}\sum\nolimits^n_{i=1}\cos 2\pi z_i)\\ + 20 + e\end{aligned}$ & $\mathbf{z} = \mathbf{M}(\mathbf{x} - \mathbf{o}) \times 0.32$ & Shifted Rotated Ackley's Function \\
				$f_{13}$ & $\begin{aligned}418.9829D-\sum\nolimits^D_{i=1}z_i\sin\sqrt{|z_i|}\end{aligned}$ & $\mathbf{z} = \mathbf{x} \times 5$ & Schwefel's Problem 2.26\\
				$f_{14}$ & $\begin{aligned}418.9829D-\sum\nolimits^D_{i=1}z_i\sin\sqrt{|z_i|}\end{aligned}$ & $\mathbf{z} = \mathbf{M}\mathbf{x} \times 5$ & Rotated Schwefel's Problem 2.26\\
				$f_{15}$ & $\begin{aligned}\sum\nolimits^D_{i=1}(z^2_i - 10\cos 2\pi z_i + 10)\end{aligned}$ & $\mathbf{z} = (\mathbf{x} - \mathbf{o}) \times 0.0512$ & Shifted Rastrigin's Function\\
				$f_{16}$ & $\begin{aligned}\sum\nolimits^D_{i=1}(z^2_i - 10\cos 2\pi z_i + 10)\end{aligned}$ & $\mathbf{z} = \mathbf{M}(\mathbf{x} - \mathbf{o}) \times 0.0512$ & Shifted Rotated Rastrigin's Function\\
				$f_{17}$ & $\begin{aligned}\sum\nolimits^D_{i=1}\frac{z^2_i}{4000}-\prod^D_{i=1}\cos\frac{z_i}{i}+1 \end{aligned}$ & $\mathbf{z} = (\mathbf{x} - \mathbf{o}) \times 6$ & Shifted Griewank's Function\\
				$f_{18}$ & $\begin{aligned}\sum\nolimits^D_{i=1}\frac{z^2_i}{4000}-\prod^D_{i=1}\cos\frac{z_i}{i}+1 \end{aligned}$ & $\mathbf{z} = \mathbf{M}(\mathbf{x} - \mathbf{o}) \times 6$ & Shifted Rotated Griewank's Function\\
				$f_{19}$ & $\begin{aligned}\sin^2(\pi y_1) + \sum\nolimits_{i=1}^{n-1}[(y_i-1)^2(1+10(\sin^2y_{i+1}))] + \\(y_n-1)^2(1+\sin^2(2\pi y_n)), y_i = 1+\frac{1}{4}(z_i+1) \end{aligned}$ & $\mathbf{z} = (\mathbf{x} - \mathbf{o}) \times 0.1$ & Shifted Levy's Function\\
				$f_{20}$ & $\begin{aligned}\sin^2(\pi y_1) + \sum\nolimits_{i=1}^{n-1}[(y_i-1)^2(1+10(\sin^2y_{i+1}))] + \\(y_n-1)^2(1+\sin^2(2\pi y_n)), y_i = 1+\frac{1}{4}(z_i+1) \end{aligned}$ & $\mathbf{z} = \mathbf{M}(\mathbf{x} - \mathbf{o}) \times 0.1$ & Shifted Rotated Levy's Function\\
				$f_{21}$\tnote{+} & $\begin{aligned}\frac{1}{10}[\sin^2(3\pi z_1)+\sum\nolimits_{i=1}^{n-1}(z_i-1)^2(1+\sin^2(3\pi z_{i+1}))+\\(z_n-1)^2(1+\sin^2(2\pi z_n))]+\sum\nolimits_{i=1}^n u(z_i,5,100,4) \end{aligned}$ & $\mathbf{z} = (\mathbf{x} - \mathbf{o}) \times 0.5$ & Shifted Penalized Function 1\\
				$f_{22}$\tnote{+} & $\begin{aligned}\frac{1}{10}[\sin^2(3\pi z_1)+\sum\nolimits_{i=1}^{n-1}(z_i-1)^2(1+\sin^2(3\pi z_{i+1}))+\\(z_n-1)^2(1+\sin^2(2\pi z_n))]+\sum\nolimits_{i=1}^n u(z_i,5,100,4) \end{aligned}$ & $\mathbf{z} = \mathbf{M}(\mathbf{x} - \mathbf{o}) \times 0.5$ & Shifted Rotated Penalized Function 1\\
				$f_{23}$\tnote{+} & $\begin{aligned}\frac{\pi}{n}[10\sin^2(\pi y_1)+\sum\nolimits_{i=1}^{n-1}(y_i-1)^2(1+10\sin^2(\pi y_{i+1}))+\\(y_n-1)^2]+\sum\nolimits_{i=1}^n u(z_i,10,100,4), y_i = 1+\frac{1}{4}(z_i+1)\end{aligned}$ & $\mathbf{z} = (\mathbf{x} - \mathbf{o}) \times 0.5$ & Shifted Penalized Function 2\\
				$f_{24}$\tnote{+} & $\begin{aligned}\frac{\pi}{n}[10\sin^2(\pi y_1)+\sum\nolimits_{i=1}^{n-1}(y_i-1)^2(1+10\sin^2(\pi y_{i+1}))+\\(y_n-1)^2]+\sum\nolimits_{i=1}^n u(z_i,10,100,4), y_i = 1+\frac{1}{4}(z_i+1)\end{aligned}$ & $\mathbf{z} = \mathbf{M}(\mathbf{x} - \mathbf{o}) \times 0.5$ & Shifted Rotated Penalized Function 2\\
				\hline			
			\end{tabular}
			\begin{tablenotes}\footnotesize
				\item [*] $\boldsymbol{o}$ is a shifting vector and $\boldsymbol{M}$ is a transformation matrix. $\boldsymbol{o}$ and $\boldsymbol{M}$ can be obtained from \cite{LiangQuSuganthanHernandez-Diaz2013ProblemDefinitionsand}.
				\item [+] $u(x,a,k,m)=\begin{cases}k(x-a)^m & \mathrm{for\:}x>a\\0& \mathrm{for\:}-a\leq x\leq a\\k(-x-a)^m & \mathrm{for\:}x<-a \end{cases}$.
			\end{tablenotes}
		\end{threeparttable}
	\end{center}
\end{table*}

In order to compare the performance improvement of our proposed ACRO algorithm over the canonical CRO in solving global optimization problems, we conduct a series of simulations on a full set of different benchmark functions. As revealed in \cite{QinHuangSuganthan2009DifferentialEvolutionAlgorithm}, many benchmark functions commonly employed for evaluating the optimization performance of evolutionary algorithms suffer from two major problems: the global optimal points are located at the center of the search space, and they are positioned along the coordinate axes, i.e., there is no correlation among different dimensions. We can shift and rotate the conventional benchmark functions to resolve these problems. We make a comprehensive test suite by employing 26 benchmark functions mainly selected from the latest CEC 2013 Real-Parameter Single Objective Optimization Competition \cite{LiangQuSuganthanHernandez-Diaz2013ProblemDefinitionsand} as well as the benchmark functions employed to test the performance of canonical CRO in \cite{LamLiYu2012RealCodedChemical}. All of these benchmark functions except $f_{13}$ and $f_{14}$ are shifted and a part of them are further rotated. The benchmark functions are listed in Table \ref{tbl:benchmark}. All benchmark functions are simulated on 30 dimensions. The search space and population initialization range are both defined to be $[-100,100]$. The global optimum value of these benchmark functions are all 0, and all simulation results smaller than $10^{-8}$ are considered as 0 \cite{LiangQuSuganthanHernandez-Diaz2013ProblemDefinitionsand}.

Both ACRO and canonical CRO are implemented in Python 2.7 on Microsoft Windows 7. All simulations are performed on a computer with an Intel Core i7-3770 @ 3.4GHz CPU. In order to reduce statistical errors and generate statistically significant results, each benchmark function is repeated for 51 independent runs for each algorithm, which also satisfies the requirement of \cite{LiangQuSuganthanHernandez-Diaz2013ProblemDefinitionsand}. In each run, we use the maximum number of function evaluations (maxFEs) as the termination criteria: 100 000 maxFEs are employed for 10 dimensional tests and 300 000 maxFEs for 30 dimensional tests \cite{LiangQuSuganthanHernandez-Diaz2013ProblemDefinitionsand}.

To evaluate the performance improvement of our proposed ACRO algorithm over the canonical CRO, we compare the simulation results between these two algorithms. Lam \textit{et al.} proposed four different CRO in \cite{LamLiYu2012RealCodedChemical}, distinguished by different constraint handling schemes, synthesis operators, and deterministic \textit{stepsize} adaptation schemes. We use CRO/BP, CRO/HP, CRO/BB, and CRO/D to refer to RCCRO1, RCCRO2, RCCRO3, and RCCRO4 in \cite{LamLiYu2012RealCodedChemical}, respectively. Similarly, we consider the two different constraint handling schemes and two synthesis operators discussed in \cite{LamLiYu2012RealCodedChemical}, and propose three variants of ACRO for performance comparison:
\begin{enumerate}
\item ACRO/BP: This ACRO variant employs the boundary constraint handling rule (9) discussed in \cite{LamLiYu2012RealCodedChemical}, and the ``probablistic select'' synthesis operator.
\item ACRO/HP: This ACRO variant is based on ACRO/BP and the boundary constraint handling rule is the hybrid handling rule (10) in \cite{LamLiYu2012RealCodedChemical}.
\item ACRO/BB: This ACRO variant is based on ACRO/BP and the synthesis operator is replaced by BLX-0.5 in \cite{LamLiYu2012RealCodedChemical}.
\end{enumerate}

The parameters for CRO are set according to the recommendation of \cite{LamLiYu2012RealCodedChemical} for solving multimodal problems, i.e. population size is 20, \textit{stepsize} is 1, initial energy buffer is $10^5$, initial kinetic energy for molecules is $10^7$, molecular collision rate is 0.2, kinetic energy loss rate is 0.1, decomposition threshold is $1.5\times10^5$, and synthesis threshold is 10. The \textit{stepsize} adaptation interval for CRO/D is set to 100 and the change rate is 0.99. As we only have three parameters for ACRO, we set the population size and the molecular collision rate the same with CRO configurations, and \textit{ChangeRate} as $10^{-4}$.

\section{Simulation Results} \label{sec:result}

\begin{table*}
	\caption{The Mean Simulation Results of ACRO and Canonical CRO}
	\label{tbl:30dresult}
	\begin{center}
		\scriptsize
		\begin{tabular}{l|lll|llll}
			\hline
			Problem & ACRO/BP & ACRO/HP & ACRO/BB & CRO/BP & CRO/HP & CRO/BB & CRO/D \\
			\hline
			$f_{1}$ & \textbf{0.0000e+00} & \textbf{0.0000e+00} & \textbf{0.0000e+00} & 2.7374e-06 & 2.6684e-06 & 2.6009e-06 & 2.9454e-05 \\
			$f_{2}$ & \textbf{0.0000e+00} & \textbf{0.0000e+00} & \textbf{0.0000e+00} & 1.8280e-05 & 1.9248e-05 & 1.8690e-05 & 1.4972e-04 \\
			$f_{3}$ & 1.9827e-06 & 5.8651e-06 & \textbf{1.4231e-06} & 4.3676e-04 & 5.3923e-04 & 4.1359e-04 & 1.4130e-01 \\
			$f_{4}$ & 6.0942e-01 & 7.3579e-01 & 1.0329e+00 & 4.7625e-03 & 4.7484e-03 & \textbf{4.4301e-03} & 1.2716e-02 \\
			$f_{5}$ & \textbf{4.8004e+00} & 7.1049e+00 & 6.1221e+00 & 5.8131e+01 & 5.8909e+01 & 5.3928e+01 & 3.5403e+01 \\
			$f_{6}$ & 2.4722e-07 & \textbf{2.1478e-07} & 9.0862e-05 & 5.2846e-04 & 5.9767e-04 & 6.2956e-04 & 1.7779e-03 \\
			$f_{7}$ & 8.4892e+01 & 1.0700e+02 & 8.7758e+01 & 1.2944e+07 & 6.8512e+06 & 4.6210e+06 & \textbf{2.2111e+01} \\
			$f_{8}$ & 2.4860e+01 & 2.9761e+01 & \textbf{2.4685e+01} & 8.4063e+01 & 8.4381e+01 & 7.1735e+01 & 6.8128e+01 \\
			$f_{9}$ & \textbf{2.2388e+01} & 2.3080e+01 & 2.2836e+01 & 3.1328e+01 & 2.9001e+01 & 2.8922e+01 & 3.0114e+01 \\
			$f_{10}$ & \textbf{1.7196e-07} & 3.7136e-01 & 4.2115e-06 & 5.1422e-03 & 5.6744e-03 & 1.1732e-02 & 1.9617e-01 \\
			$f_{11}$ & 8.4478e-02 & 9.2534e-02 & 7.8069e-02 & 1.1503e+01 & 1.1305e+01 & 1.1177e+01 & \textbf{2.1326e-04} \\
			$f_{12}$ & \textbf{2.3127e+00} & 2.4233e+00 & 2.3163e+00 & 1.1388e+01 & 1.1562e+01 & 1.1231e+01 & 3.0470e+00 \\
			$f_{13}$ & 2.9281e+02 & 6.8552e+02 & 2.3939e+02 & 5.8231e+03 & 6.0672e+03 & 6.1206e+03 & \textbf{0.0000e+00} \\
			$f_{14}$ & 4.1386e+03 & 3.6792e+03 & 3.8300e+03 & 5.7527e+03 & 5.5973e+03 & 5.6308e+03 & \textbf{8.6744e+02} \\
			$f_{15}$ & 6.9226e+00 & 6.1607e+00 & 4.8700e+00 & 4.2171e+02 & 4.3384e+02 & 3.8941e+02 & \textbf{1.4585e-04} \\
			$f_{16}$ & 1.6828e+02 & 1.9685e+02 & \textbf{1.6156e+02} & 4.3566e+02 & 4.5059e+02 & 4.1632e+02 & 2.4930e+02 \\
			$f_{17}$ & 3.0256e-03 & 2.3141e-03 & \textbf{1.5916e-03} & 2.3064e+00 & 2.1902e+00 & 2.6168e+00 & 1.0624e-02 \\
			$f_{18}$ & 5.8008e-04 & \textbf{1.9328e-04} & 2.9004e-04 & 4.7648e-03 & 1.9807e-03 & 4.0542e-03 & 9.1952e-03 \\
			$f_{19}$ & \textbf{0.0000e+00} & \textbf{0.0000e+00} & \textbf{0.0000e+00} & 1.9428e+01 & 1.2036e+01 & 1.0251e+01 & 1.8466e-08 \\
			$f_{20}$ & 3.9330e+01 & 4.3081e+01 & 4.0080e+01 & \textbf{3.7536e+01} & 3.8191e+01 & 6.9319e+01 & 7.2789e+01 \\
			$f_{21}$ & 4.0374e-08 & 3.4969e-06 & \textbf{9.4729e-10} & 2.6110e-07 & 2.3093e-07 & 2.1121e-07 & 1.5477e-06 \\
			$f_{22}$ & 2.2444e+00 & 2.2262e+00 & 2.5946e+00 & 2.9582e+01 & 2.7701e+01 & 2.8518e+01 & \textbf{1.9143e+00} \\
			$f_{23}$ & 1.8949e-03 & 1.7269e-05 & 1.2809e-03 & 1.6633e+01 & 1.5481e+01 & 6.9264e+00 & \textbf{1.5812e-07} \\
			$f_{24}$ & 3.5586e+00 & 3.5509e+00 & 3.4288e+00 & \textbf{2.1111e+00} & 2.4530e+00 & 2.5946e+00 & 3.1687e+00 \\
			\hline
		\end{tabular}
	\end{center}
\end{table*}

In this section we present the simulation results of different variants of our proposed ACRO and canonical CRO described in Section \ref{sec:experiment}. We present the raw simulation result data as well as the statistical analysis based on the data. We also analyze the convergence of the different algorithms.

The mean value obtained by all ACRO and canonical CRO algorithms on different benchmark functions are presented in Table \ref{tbl:30dresult}. The mean values in bold font indicate superiority of the corresponding algorithm over the benchmark function. The simulation results indicate that ACRO generally gives superior performance over the canonical CRO algorithms. ACRO algorithms generate best results in 14 out of 24 30-D functions.

From the simulation results in Table \ref{tbl:30dresult} the following key points can be observed:

\begin{enumerate}
\item ACRO algorithms generally have better performance than CRO algorithms in all benchmark functions when comparing the mean simulation results, especially in uni-modal functions ($f_1$--$f_{10}$). Meanwhile, the superiority of ACRO in solving multi-modal functions is still very significant. While the deterministic $stepSize$ adaptation scheme proposed in \cite{LamLiYu2012RealCodedChemical} has better overall performance than other CRO variants, ACRO algorithms outperform it by a large margin.

\item While CRO has no significant preference on the boundary handling scheme selection, ACRO favors the boundary constraint handling rule (9) discussed in \cite{LamLiYu2012RealCodedChemical} more than the hybrid handling rule discussed in the same paper. This phenomenon can be observed in the benchmark functions where ACRO/BP performs better than ACRO/HP, e.g. $f_7$ and $f_{13}$.

\item ACRO has no significant preference between the two different types of synthesis employed by ACRO/BB and ACRO/BP. These two algorithms perform similarly in all the benchmark functions tested.
\end{enumerate}

In terms of the computational time consumed by the proposed adaptation scheme, we observed that the additional time needed for ACRO algorithms is less than 5\% of the computation time of canonical CRO algorithms. As the computational overhead of the evolutionary algorithms on a given number of function evaluations with the same parameter setting is approximately constant when solving different optimization problems, we believe that the extra time used by the adaptation scheme can be considered negligible when solving problems with relatively long fitness evaluation time.

\section{Conclusion} \label{sec:conclusion}

In this paper, we proposed a new CRO-based metaheuristic called ACRO. To alleviate the parameter tuning process, we reduce the number of parameters defined in CRO. We also design an adaptive scheme for CRO. We manipulate six of the existing CRO parameters and successfully reduce the total number of parameters from eight to three. The advantages of ACRO over canonical CRO include:
\begin{enumerate}
\item ACRO achieves superior optimization performance over canonical CRO within a similar amount of time.

\item ACRO convergences faster than canonical CRO, and can effectively prevent pre-mature convergence.

\item ACRO can significantly reduce the time needed to find a suitable combination of optimization parameters.
\end{enumerate}
These advantages can be observed in the simulation we conducted over a set of 24 benchmark functions. The simulation results demonstrate the superiority of ACRO.

In the future, we will test the performance of ACRO on other practical problems. We will refine the settings for the proposed adaptive schemes. We can also include ACRO in an ensemble-based algorithm for real-parameter optimization \cite{MallipeddiMallipeddiSuganthanTasgetiren2011DifferentialEvolutionAlgorithm}\cite{MallipeddiSuganthan2010EnsembleConstraintHandling}. Developing ACRO-hybrid algorithms with other swarm intelligence algorithms, such as Particle Swarm Optimization \cite{KennedyEberhart1995Particleswarmoptimization} and Social Spider Algorithm \cite{YuLi2015SocialSpiderAlgorithm}, is also a potential research topic.

\bibliographystyle{IEEEtran}
\bibliography{IEEEabrv,../../../../bib/publications}

\end{document}